\documentclass[conference]{IEEEtran}
\IEEEoverridecommandlockouts

\usepackage{cite}
\usepackage{amsmath,amssymb}
\usepackage{graphicx}
\usepackage{textcomp}
\usepackage{xcolor}
\usepackage{booktabs}
\usepackage{multirow}
\usepackage{array}
\usepackage{url}
\usepackage{balance}

\usepackage{dblfloatfix}

\usepackage{placeins}

\def\BibTeX{{\rm B\kern-.05em{\sc i\kern-.025em b}\kern-.08em
    T\kern-.1667em\lower.7ex\hbox{E}\kern-.125emX}}

\begin{document}

\title{Quantifying the Affective Gap: A Zero-Shot Evaluation of LLMs on
Fine-Grained Emotion Taxonomies}

\author{ \IEEEauthorblockN{ Lawrence Obiuwevwi, Krzysztof J. Rechowicz, Jessica M. Johnson, \\ Vikas Ashok, Sachin Shetty, \& Sampath Jayarathna } \IEEEauthorblockA{ \textit{Old Dominion University, Norfolk, VA, USA} \\ lobiu001@odu.edu, krechowi@odu.edu, j17johnso@odu.edu, \\ vganjiqu@cs.odu.edu, sshetty@odu.edu, sampath@cs.odu.edu } }

\maketitle

\begin{abstract}
Emotion recognition in natural language is a foundational challenge in
affective computing, with critical implications for human-computer
interaction, mental health support, and conversational AI. This paper
presents a rigorous, unified zero-shot evaluation of three leading commercial
large language models: Claude (claude-sonnet-4-6), ChatGPT (GPT-5.4), and
Gemini (gemini-2.5-flash). The models were queried through their respective
production APIs as of April 2026 on a fine-grained 13-class emotion
classification task. Using a stratified 1,000-sentence sample from the
\textit{boltuix/emotions-dataset} comprising 131,306 sentences across
13 categories, a single uniform prompt with no exemplars was applied
identically across all models. Gemini achieves the highest accuracy
(39.9\%) and macro-F1 (0.363), followed by GPT-5.4 (38.8\%,
F1\,=\,0.291) and Claude (38.0\%, F1\,=\,0.159). All models excel on
\textit{sarcasm} and \textit{desire} while consistently failing on
\textit{love}, \textit{confusion}, and \textit{shame}. McNemar's tests
reveal no statistically significant pairwise differences ($p>0.10$),
suggesting convergence at a shared zero-shot ceiling. Claude's markedly
lower macro-F1 exposes a class-imbalance prediction bias. These findings
highlight the current limitations of frontier AI systems in zero-shot
fine-grained emotion classification.
\end{abstract}

\begin{IEEEkeywords}
emotion classification, affective computing, LLMs, zero-shot learning, NLP,
GPT-5.4, Claude, Gemini, AI, human-computer interaction
\end{IEEEkeywords}

\section{Introduction}
\label{sec:intro}

Emotions shape human communication and underpin high-stakes applications
including mental health monitoring~\cite{chancellor2020methods}, empathetic
dialogue systems~\cite{rashkin2019towards}, and human-computer interaction.
The ability of an AI system to recognize affective states is therefore a
prerequisite for deployment in any domain where human welfare is at stake.

Despite rapid advances in large language models (LLMs), their affective
intelligence remains poorly characterized at fine-grained resolution. Most
studies probe coarse polarity or the Ekman six basic
emotions~\cite{ekman1992argument} rather than richer
taxonomies~\cite{yang2023evaluating,kheiri2023sentimentgpt}, and direct
cross-provider comparisons under identical experimental conditions are absent
from the literature. This gap matters: practitioners selecting an API for
emotion-aware applications must rely on ad hoc or proprietary benchmarks that
may not reflect real-world linguistic diversity.

This paper addresses the gap through four research questions: zero-shot
accuracy across 13 classes (RQ1); pairwise statistical differences using
McNemar's test (RQ2); per-class strengths and failure modes (RQ3); and the
effect of sentence length on accuracy (RQ4). We contribute \textit{(i)} the
first direct zero-shot comparison of Claude (claude-sonnet-4-6), ChatGPT
(GPT-5.4), and Gemini (gemini-2.5-flash) on a 13-class task using production
APIs as of April 2026; \textit{(ii)} per-emotion accuracy breakdowns revealing
systematic failure modes; \textit{(iii)} McNemar significance testing; and
\textit{(iv)} a sentence-length moderation analysis.

\section{Related Work}
\label{sec:related}

\subsection{Emotion Classification and Datasets}

Computational emotion recognition has progressed from lexicon-based resources
such as the NRC Lexicon~\cite{mohammad2013crowdsourcing} and SemEval affective
tasks~\cite{strapparava2007semeval,mohammad2018semeval,chatterjee2019semeval2019task3}
through deep learning classifiers to transformer fine-tuning paradigms
achieving state-of-the-art results on benchmarks including
GoEmotions~\cite{demszky2020goemotions} (58K comments, 27 categories).
BERT~\cite{devlin2019bert,pineda2023using} and RoBERTa~\cite{liu2019roberta} established the
transformer paradigm as dominant for supervised emotion classification.
The \textit{boltuix/emotions-dataset} extends the Ekman taxonomy with social
(shame, guilt)~\cite{tangney2002shame}, cognitive (confusion), rhetorical
(sarcasm)~\cite{joshi2017automatic}, and motivational (desire) categories,
offering broader ecological validity.

Emotion-aware AI may also complement multimodal assistive systems that integrate cognitive, physiological, and attentional context in knowledge-work environments~\cite{obiuwevwi2026cognitive}. Related human-computer interaction research has examined how eye-tracking signals can characterize variations in cognitive effort during digital-document reading~\cite{mahanama2021analyzing, mahanama2022multidisciplinary, thennakoon2025devices}.

\subsection{LLMs for Emotion and Sentiment}

GPT-3~\cite{brown2020language} demonstrated zero-shot classification through
in-context learning, with instruction
tuning~\cite{ouyang2022training,wei2021finetuned} substantially improving
generalization. SentimentGPT~\cite{kheiri2023sentimentgpt} found LLMs
competitive on coarse polarity but weaker on low-frequency emotion classes.
Yang et al.~\cite{yang2023evaluating} evaluated ChatGPT and GPT-4 on
sentiment tasks but excluded Claude, Gemini, and a fine-grained 13-class
taxonomythe gap this study directly addresses.

\section{Methodology}
\label{sec:methodology}

\subsection{Dataset}

The \textit{boltuix/emotions-dataset}~\cite{boltuix2023emotions} is a
publicly available Hugging Face corpus of 131,306 English sentences annotated
with one of 13 mutually exclusive emotion labels: \textit{happiness, sadness,
fear, anger, love, disgust, surprise, neutral, confusion, desire, shame,
guilt,} and \textit{sarcasm}. A stratified random sample of $N=1{,}000$
sentences was drawn with seed\,=\,42, preserving class proportions to
simulate realistic deployment conditions. Table~\ref{tab:class_dist} reports
the distribution. \textit{Happiness} is the most frequent class (20.5\%,
$n=205$); \textit{desire} is the rarest (1.9\%, $n=19$). This 10:1 ratio has
direct consequences for macro-F1: models concentrating predictions on
high-frequency classes achieve deceptively high accuracy while failing on
minority classes.

\begin{table}[!t]
  \caption{Class Distribution in Stratified Sample ($N=1{,}000$)}
  \label{tab:class_dist}
  \centering
  \small
  \begin{tabular}{lrr lrr}
    \toprule
    \textbf{Emotion} & \textbf{Count} & \textbf{\%} &
    \textbf{Emotion} & \textbf{Count} & \textbf{\%} \\
    \midrule
    happiness  & 205 & 20.5 & disgust   &  60 &  6.0 \\
    neutral    & 135 & 13.5 & surprise  &  44 &  4.4 \\
    sadness    & 130 & 13.0 & shame     &  34 &  3.4 \\
    anger      & 107 & 10.7 & sarcasm   &  26 &  2.6 \\
    love       &  76 &  7.6 & guilt     &  26 &  2.6 \\
    confusion  &  70 &  7.0 & desire    &  19 &  1.9 \\
    fear       &  68 &  6.8 &           &     &      \\
    \midrule
    \multicolumn{6}{r}{\textbf{Total: 1,000 (100.0\%)}} \\
    \bottomrule
  \end{tabular}
\end{table}

\subsection{Models and API Configuration}

Three frontier LLMs were queried through their production APIs as of April 2,
2026: Claude (claude-sonnet-4-6) through Anthropic's Messages API with default
sampling; ChatGPT (GPT-5.4) through OpenAI's Chat Completions API with
\texttt{temperature=0}; and Gemini (gemini-2.5-flash) through Google's
Generative AI API with default sampling. All models received
\texttt{max\_tokens=10}, enforcing single-word responses. Responses not
matching one of the 13 canonical labels after case normalization were treated
as errors; all three models exhibited strong label adherence in practice.

\subsection{Prompt Design}

A single-turn prompt was used identically across all models, with no system
message and no few-shot examples~\cite{brown2020language,wei2022chain}:

\smallskip
\noindent\small\textit{``Classify the emotion expressed in this sentence.
Reply with ONLY one word from this list: happiness, sadness, fear, anger,
love, disgust, surprise, neutral, confusion, desire, shame, guilt,
sarcasm. Sentence: \{sentence\}. Emotion:''}
\smallskip

\normalsize
The full label list constrains the output space; imperative instruction style
elicits consistent outputs from instruction-tuned
models~\cite{ouyang2022training}. No chain-of-thought preamble is included,
ensuring performance reflects raw affective inference~\cite{wei2022chain}.

\subsection{Evaluation Metrics}

Four metrics were computed per model: (i)~\textbf{Overall Accuracy};
(ii)~\textbf{Macro-F1} ($F1_\text{mac}$), the primary metric for
minority-class evaluation; (iii)~\textbf{Weighted-F1} ($F1_\text{wt}$); and
(iv)~\textbf{Cohen's Kappa} ($\kappa$)~\cite{cohen1960coefficient}.
Pairwise comparisons used \textbf{McNemar's test}~\cite{mcnemar1947note} with
continuity correction.

\section{Results}
\label{sec:results}

\subsection{Overall Performance}

Table~\ref{tab:overall} summarizes aggregate performance. Gemini leads on all
four metrics (accuracy 39.9\%, macro-F1 0.363, $\kappa=0.337$), followed by
GPT-5.4 (38.8\%, 0.291, 0.327) and Claude (38.0\%, 0.159, 0.320). All three
models substantially exceed the majority-class baseline (20.5\%) and random
baseline (7.7\%), yet fall far below human-level performance
($>$70\%)~\cite{demszky2020goemotions}. The narrow 1.9\,pp spread suggests
convergence near a shared zero-shot ceiling, confirmed by the McNemar results
below.

\begin{table}[!t]
  \caption{Overall Performance ($N=1{,}000$)}
  \label{tab:overall}
  \centering
  \small
  \begin{tabular}{lcccc}
    \toprule
    \textbf{Model} & \textbf{Acc.\ (\%)} & \textbf{F1$_\text{mac}$} &
    \textbf{F1$_\text{wt}$} & \textbf{$\kappa$} \\
    \midrule
    Claude (sonnet-4-6) & 38.0 & 0.159 & 0.394 & 0.320 \\
    GPT-5.4             & 38.8 & 0.291 & 0.387 & 0.327 \\
    Gemini (2.5-flash)  & \textbf{39.9} & \textbf{0.363} &
                          \textbf{0.394} & \textbf{0.337} \\
    \bottomrule
  \end{tabular}
\end{table}

\subsection{Per-Emotion Accuracy}

\begin{figure}[!t]
  \centering
  \includegraphics[width=\linewidth]{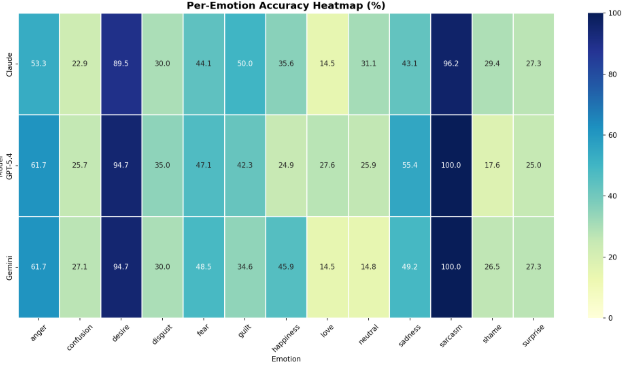}
  \caption{Per-emotion accuracy heatmap (\%). Sarcasm scores highest
  (96.2--100\%); love and neutral are consistently weakest.}
  \label{fig:heatmap}
\end{figure}

Table~\ref{tab:per_emotion} and Fig.~\ref{fig:heatmap} present per-class
accuracy. \textit{Sarcasm} (96.2\%--100\%) and \textit{desire}
(89.5\%--94.7\%) stand in stark contrast to all other categories.
\textit{Anger} (53.3\%--61.7\%), \textit{fear} (44.1\%--48.5\%), and
\textit{sadness} (43.1\%--55.4\%) form a mid-performance cluster.
\textit{Love} (14.5\%--27.6\%), \textit{confusion} (22.9\%--27.1\%), and
\textit{shame} (17.6\%--29.4\%) are consistently the hardest categories.
Notable model-specific divergences include Gemini collapsing on
\textit{neutral} (14.8\% vs.\ Claude's 31.1\%), Claude leading on
\textit{guilt} (50.0\%) and \textit{neutral}, and GPT-5.4 leading on
\textit{sadness} and \textit{love}.

\begin{table}[!t]
  \caption{Per-Emotion Accuracy (\%) by Model}
  \label{tab:per_emotion}
  \centering
  \small
  \begin{tabular}{lrrr}
    \toprule
    \textbf{Emotion} & \textbf{Claude} & \textbf{GPT-5.4} &
    \textbf{Gemini} \\
    \midrule
    sarcasm    & 96.2          & \textbf{100.0} & \textbf{100.0} \\
    desire     & 89.5          & \textbf{94.7}  & \textbf{94.7}  \\
    anger      & 53.3          & 61.7           & 61.7           \\
    guilt      & \textbf{50.0} & 42.3           & 34.6           \\
    fear       & 44.1          & 47.1           & \textbf{48.5}  \\
    sadness    & 43.1          & \textbf{55.4}  & 49.2           \\
    happiness  & 35.6          & 24.9           & \textbf{45.9}  \\
    disgust    & 30.0          & \textbf{35.0}  & 30.0           \\
    neutral    & \textbf{31.1} & 25.9           & 14.8           \\
    shame      & \textbf{29.4} & 17.6           & 26.5           \\
    surprise   & 27.3          & 25.0           & 27.3           \\
    confusion  & 22.9          & 25.7           & \textbf{27.1}  \\
    love       & 14.5          & \textbf{27.6}  & 14.5           \\
    \midrule
    \textbf{Overall} & 38.0    & 38.8           & \textbf{39.9}  \\
    \bottomrule
  \end{tabular}
\end{table}

\begin{figure*}[!t]
  \centering
  \includegraphics[width=\textwidth]{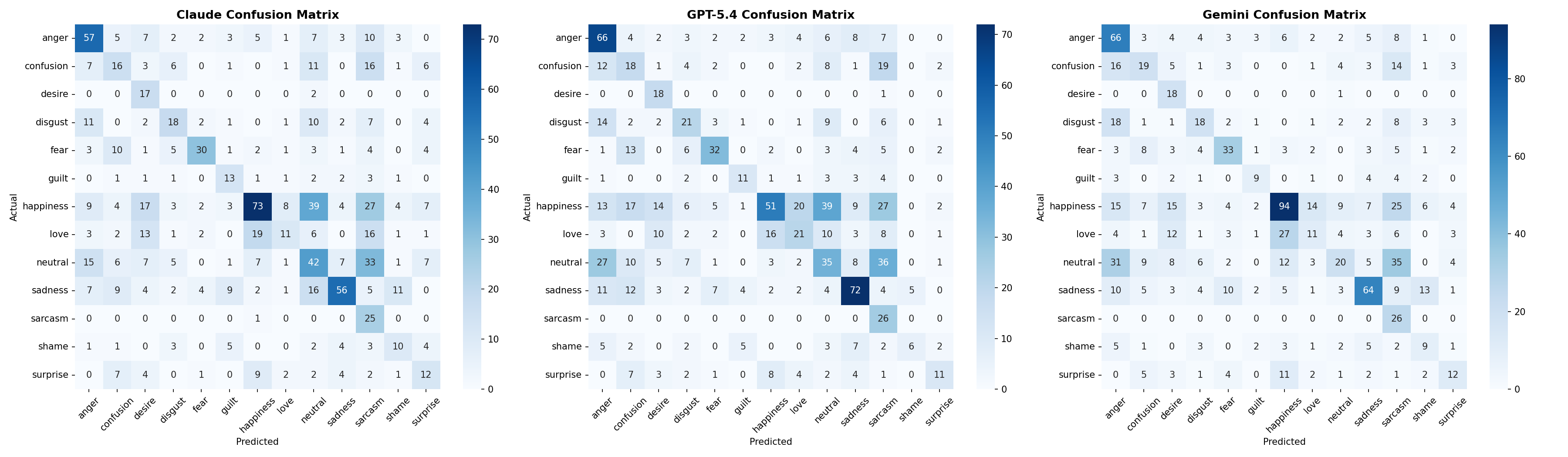}
  \caption{Normalized confusion matrices for Claude, GPT-5.4, and Gemini.
  Off-diagonal mass concentrates on semantically proximal pairs:
  \textit{love}$\leftrightarrow$\textit{happiness},
  \textit{fear}$\leftrightarrow$\textit{sadness}, and
  \textit{confusion}$\leftrightarrow$\textit{neutral}.}
  \label{fig:confusion}
\end{figure*}

Fig.~\ref{fig:confusion} presents normalized confusion matrices. Cross-model
patterns include \textit{love} misclassified as \textit{happiness}, confirming
semantic overlap between positive affect categories; \textit{confusion}
misclassified as \textit{neutral} and \textit{sadness}; and \textit{fear}
frequently assigned to \textit{sadness}, suggesting that shared negative
valence dominates arousal-level distinctions.


\subsection{Statistical Significance}

Table~\ref{tab:mcnemar} presents McNemar's test $p$-values. No comparison
reaches significance at $\alpha=0.05$, supporting convergence: accuracy
differences are consistent with random variation, and no model can be declared
definitively superior under this protocol.

\begin{table}[!t]
  \caption{McNemar's Test $p$-Values ($\alpha=0.05$)}
  \label{tab:mcnemar}
  \centering
  \small
  \begin{tabular}{lcc}
    \toprule
    \textbf{Comparison} & \textbf{$p$-value} & \textbf{Significant?} \\
    \midrule
    Claude vs.\ GPT-5.4  & 0.5083 & No \\
    Claude vs.\ Gemini   & 0.1350 & No \\
    GPT-5.4 vs.\ Gemini  & 0.4095 & No \\
    \bottomrule
  \end{tabular}
\end{table}

\subsection{Sentence Length Analysis}

\begin{figure}[!t]
  \centering
  \includegraphics[width=\linewidth]{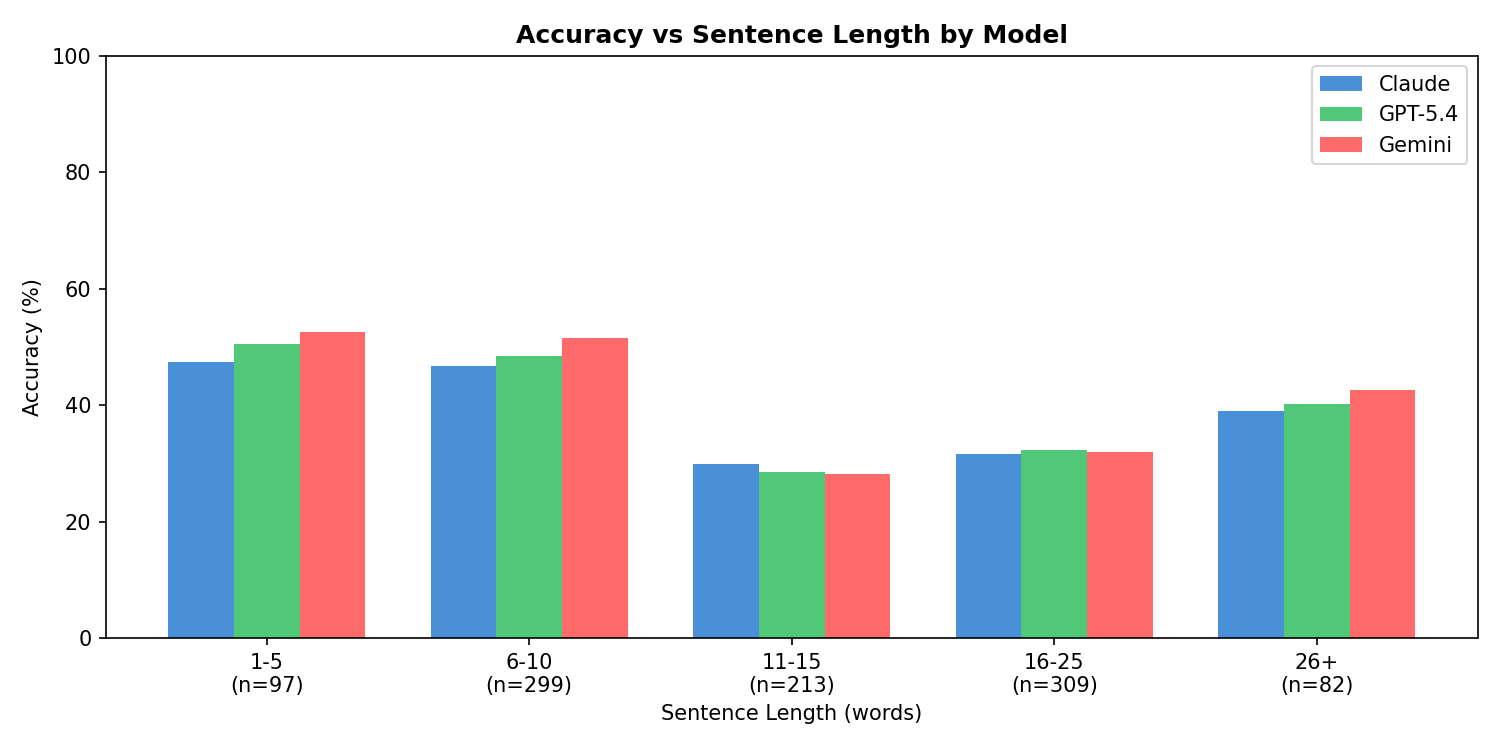}
  \caption{Accuracy by sentence length. Medium sentences (5--15 words)
  yield peak performance; long sentences ($>$15 words) degrade accuracy
  by $\sim$9\,pp across all models.}
  \label{fig:length}
\end{figure}

Performance peaks in the medium range (5--15 words): Claude\,=\,41.3\%,
GPT-5.4\,=\,42.0\%, and Gemini\,=\,43.6\% ($n=567$). Long sentences
($>$15 words) produce the lowest accuracy (33.2\%--34.3\%), a $\sim$9\,pp
drop consistent across all models (Fig.~\ref{fig:length}), suggesting a
structural property of zero-shot affective inference.


\section{Discussion}
\label{sec:discussion}

\textbf{Zero-shot ceiling:}
The 1.9\,pp accuracy spread and non-significant McNemar results indicate all
three models have converged near a shared zero-shot ceiling. For practitioners,
provider choice is unlikely to affect raw accuracy. However, the macro-F1 gap
(0.159 vs.\ 0.363, Claude vs.\ Gemini) is practically significant: in
applications requiring minority-class sensitivity (e.g., shame, guilt),
Gemini is the preferred choice.

\textbf{Sarcasm/desire paradox and hard categories:}
Near-perfect sarcasm (96.2\%--100\%) and desire (89.5\%--94.7\%) accuracies
likely reflect small support sizes ($n=26$, $n=19$) and atypical lexical
prototypicality~\cite{larochelle2008zero} rather than robust affective
understanding. Conversely, \textit{love} is the worst-performing class for
Claude and Gemini (14.5\%), driven by semantic overlap with
\textit{happiness}. The confusion matrix confirms that most \textit{love}
misclassifications land on \textit{happiness}, consistent with
GoEmotions~\cite{demszky2020goemotions}, where positive social emotions formed
the hardest discrimination cluster.

\textbf{Provider-specific biases:}
Claude's macro-F1 of 0.159less than half of Gemini's 0.363despite
comparable accuracy (38.0\%) is the signature of majority-class prediction
bias~\cite{wei2021finetuned}. Claude over-predicts \textit{happiness} while
underperforming on minority classes, likely amplified by alignment processes
during pretraining~\cite{ouyang2022training}. This is a critical limitation
for mental health applications where rare emotions carry high clinical
significance. Gemini, by contrast, collapses on \textit{neutral} (14.8\%
vs.\ Claude's 31.1\%), systematically reassigning neutral sentences to
emotional categories. This pattern may reflect RLHF objectives rewarding
emotionally engaging outputs~\cite{ouyang2022training}, making Gemini
unreliable for applications requiring neutral-state detection.

\textbf{Implications and limitations:}
Accuracy in the 38\%--40\% range falls well short of deployment thresholds
for emotionally sensitive
contexts~\cite{chancellor2020methods,rashkin2019towards}. Key limitations
include the zero-shot-only evaluation (few-shot prompting may substantially
improve all models), reliance on a single English-language dataset, and
high-variance estimates for rare classes ($n<30$). The results reflect API
behavior as of April 2, 2026; continuous model updates may alter findings.

\section{Conclusion}
\label{sec:conclusion}

We presented the first direct zero-shot evaluation of Claude, GPT-5.4, and
Gemini on 13-class emotion classification using production APIs. Gemini led
on all metrics (accuracy 39.9\%, macro-F1 0.363), although no pairwise
difference was statistically significant, suggesting convergence near a
shared zero-shot ceiling. 
Our results reveal that frontier LLMs remain substantially
below human-level affective awareness in zero-shot fine-grained settings.
Future work should explore few-shot and chain-of-thought
prompting~\cite{wei2022chain} for minority-class improvement, fine-tuned
baselines to quantify the zero-shot gap, and multilingual evaluation to probe
cross-lingual affective generalization.

 \section*{Acknowledgement}
This work is supported in part by NSF 245523. Any opinions, findings, and conclusions or recommendations expressed in this material are of the author(s) and do not necessarily reflect those of the sponsors.

\FloatBarrier

\balance
\bibliographystyle{IEEEtran}
\bibliography{emotion}

\end{document}